\documentclass[10pt,twocolumn,letterpaper]{article}

\usepackage{cvpr}      

\definecolor{veronica-red}{RGB}{196,30,58}
\newcommand{\sqs}[1]{{\color{veronica-red}{[(SQS): #1]}}}
\usepackage{xspace}
\usepackage{colortbl}

\usepackage{longtable}

\usepackage{booktabs}

\usepackage{array}

\usepackage{ragged2e}
\usepackage{setspace}

\usepackage{multicol}

\usepackage{listings}
\usepackage{courier}

\lstdefinelanguage{json}{
    basicstyle=\ttfamily\footnotesize,
    breaklines=true,
    showstringspaces=false,
}

\definecolor{cvprblue}{rgb}{0.21,0.49,0.74}

\usepackage[pagebackref,breaklinks,colorlinks,allcolors=cvprblue]{hyperref}
\usepackage[dvipsnames]{xcolor} 
\usepackage{pifont}
\usepackage{makecell}
\usepackage{multirow}
\def\paperID{} 
\def\confName{}

\title{OS-Oracle: A Comprehensive Framework for Cross-Platform GUI Critic Models}

\author{
\textbf{Zhenyu Wu}\textsuperscript{$1,2$}\thanks{\, Equal Contribution.}\;,
\textbf{Jingjing Xie}\textsuperscript{$3\,*$},
\textbf{Zehao Li}\textsuperscript{$2$}, \textbf{Bowen Yang}\textsuperscript{$2$}, \textbf{Qiushi Sun}\textsuperscript{$4$}, \textbf{Zhaoyang Liu}\textsuperscript{$5$},\\
\textbf{Zhoumianze Liu}\textsuperscript{$2$}, \textbf{Yu Qiao}\textsuperscript{$2$}, \textbf{Xiangyu Yue}\textsuperscript{$3$}, \textbf{Zun Wang}\textsuperscript{$2$}, \textbf{Zichen Ding}\textsuperscript{$2$}\thanks{\, Corresponding Author.} \\
\textsuperscript{$1$}Shanghai Jiaotong University 
\textsuperscript{$2$}Shanghai AI Laboratory 
\textsuperscript{$3$}CUHK MMLab \\
\textsuperscript{$4$}The University of Hong Kong
\textsuperscript{$5$}The Hong Kong University of Science and Technology\\
}

\newcommand{\name}{OS-Oracle}

\newcommand{\cmark}{\textcolor{green}{\ding{51}}} 
\newcommand{\xmark}{\textcolor{red}{\ding{55}}} 

\begin{document}
\maketitle
\begin{abstract}
With VLM-powered computer-using agents (CUAs) becoming increasingly capable at graphical user interface (GUI) navigation and manipulation, reliable step-level decision-making has emerged as a key bottleneck for real-world deployment. In long-horizon workflows, errors accumulate quickly and irreversible actions can cause unintended consequences, motivating critic models that assess each action before execution. While critic models offer a promising solution, their effectiveness is hindered by the lack of diverse, high-quality GUI feedback data and public critic benchmarks for step-level evaluation in computer use.
To bridge these gaps,
we introduce OS-Oracle that makes three core contributions:
(1) a scalable data pipeline for synthesizing cross-platform GUI critic data;
(2) a two-stage training paradigm combining supervised fine-tuning (SFT) and consistency-preserving group relative policy optimization (CP-GRPO); (3) OS-Critic Bench, a holistic benchmark for evaluating critic model performance across Mobile, Web, and Desktop platforms.
Leveraging this framework, we curate a high-quality dataset containing 310k critic samples. 
The resulting critic model, OS-Oracle-7B, 
achieves state-of-the-art performance among open-source VLMs on OS-Critic Bench, and surpasses proprietary models on the mobile domain. Furthermore, when serving as a pre-critic, OS-Oracle-7B improves the performance of native GUI agents such as UI-TARS-1.5-7B in OSWorld and AndroidWorld environments.
The code is open-sourced at \url{https://github.com/numbmelon/OS-Oracle}.

\end{abstract}    
\section{Introduction}
\label{sec:intro}

\begin{figure*}[ht]
  \centering
  \includegraphics[width=0.8\linewidth]{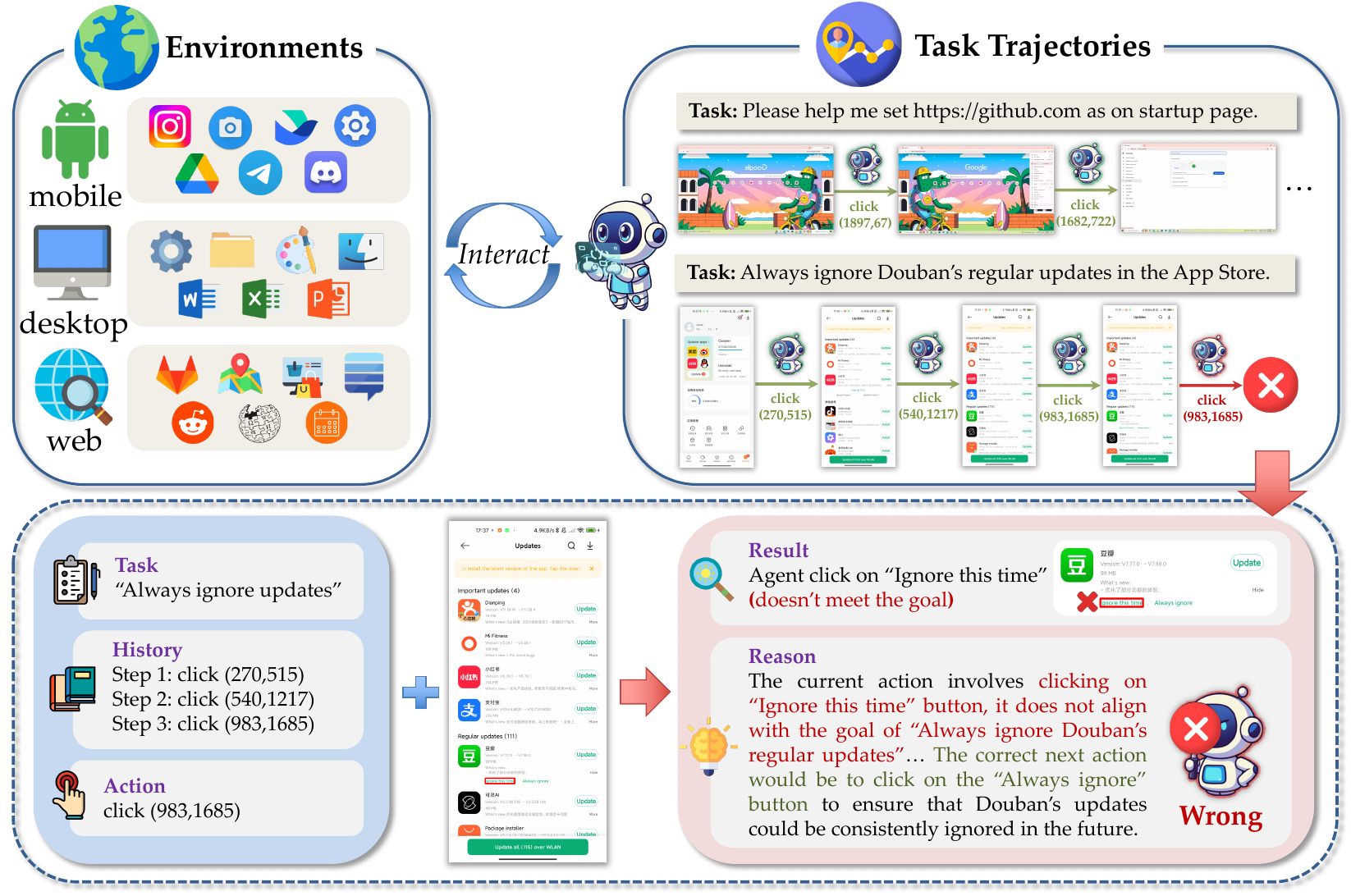}
\caption{An example of CUA automation with OS-Orcale-7B. The critic model analyzes the current task, interaction history, the CUA's proposed next action, and the current screenshot, and then judges whether the action is correct.}
  \label{fig:teaser}
\end{figure*}

Recent advances in Vision-Language Models (VLMs)~\cite{gpt_5,wang2025internvl35,bai2025qwen25vl,guo2025seed1,claude_45,vteam2025glm45vglm41vthinkingversatilemultimodal} not only demonstrate remarkable performance on general vision-language tasks, but also exhibit promising potential in understanding, navigating, and manipulating graphical user interfaces (GUIs). As such, VLM-based computer-using agents (CUAs)~\cite{hong2024cogagent,cheng-etal-2024-seeclick,wu2024atlas,qin2025uitars,liu2025scalecua,wang2024gui,yang2025zerogui,nguyen2025gui} have attracted substantial attention in community. By interacting with GUI environments (\eg, Desktop, Mobile, web, \etc), these agents are capable of completing user-specified tasks autonomously, laying a solid foundation for digital task automation.

Despite rapid progress, current CUAs still expose several recurring limitations in practice: 1) \textit{Operation Failure} (OF), which reflects a failure to perceive subtle state transitions, leading to mistimed \texttt{type} before \texttt{click}, redundant post-click retries, or futile boundary scrolling; 2) \textit{Inefficient Error State Recovery} (IESR), which arises when agents enter unexpected UI states yet fail to issue corrective operations such as \texttt{back}, allowing errors to compound; 3) \textit{Mistimed Task Termination} (MTT), which denotes an imprecise understanding of termination states, causing premature stopping on unfinished tasks or unnecessary steps after successful completion; And 4) \textit{Inaccurate Element Localization} (IEL), which often occurs when the wrong UI elements are selected or when the click position falls outside the activation area of the UI element.
The first three issues primarily stem from weaknesses in planning and step-level reasoning under partial observability and long horizons, whereas IEL reflects limited UI element perception and grounding. These deficiencies degrade robustness across desktop, mobile, and web environments and hinder reliable long-horizon task completion.

To mitigate these limitations, existing efforts follow two main routes. The first line optimizes native agents with post-training optimization based on offline or online reinforcement learning (RL)~\cite{qin2025uitars,lai2025computerrl,ye2025mobile,wang2025ui,gu2025ui}, which can improve planning, error recovery, and termination decisions, and sometimes alleviate localization errors for models. Yet it is difficult to optimize, sensitive to reward design, and costly due to large-scale environment interaction, particularly for complex tasks that involve long-horizon planning and reasoning. 
The alternative route focuses on the Critic Model~\cite{sun2025seagent,wanyan2025look,xiao2025ui}, which utilizes a verifier to assess the correctness of each step when agents interact with environments, offering a more cost-effective and scalable solution. Critic models can be trained on smaller VLMs and then applied to various agents without extensive retraining. However, despite their potential, Critic models for computer use are still underdeveloped due to the lack of i) \textit{open-source multi-platform critic data for training} and ii) \textit{a well-defined benchmark for evaluation}. 
As a result, there remains significant room to advance the development of Critic Models. 

To tackle these gaps, we present \textbf{\name}, a full-stack framework that integrates data curation, training recipes, and evaluation protocols for developing critic models. \textbf{First}, we build an automatic critic data pipeline that derives hard negatives from existing positive trajectories to target the four error types, \ie, OF, IESR, MTT, and IEL. We begin by extracting positive triplets \textless\texttt{instruction,screenshot,action}\textgreater\ from trajectory datasets. We then apply a rule-based policy to convert the gold action into a type-specific erroneous action. For each negative sample, we inject the explicit error tag into the prompt and use GPT-4o to annotate concise rationales, yielding a high-quality corpus of $\sim$160k positives and $\sim$150k negatives. \textbf{Second}, based on Qwen2.5VL-7B~\cite{bai2025qwen25vl}, we train a visual critic model termed as \textit{\name-7B} with a two-stage recipe. Stage one uses \textit{supervised fine-tuning} to establish core discrimination and rationale skills. Stage two applies \textit{Consistency-Preserving Group Relative Policy Optimization} (CP-GRPO), which utilizes a consistency reward to align the model’s reasoning content with its final judgment, improving both discriminability and reasoning–judgment agreement. \textbf{Third}, we construct OS-Critic Bench, the first cross-platform benchmark for mobile, web, and desktop. We sample trajectories from existing computer-using benchmarks, and generate one action proposal per screenshot with Qwen2.5VL-7B due to its balanced accuracy in practice. Then, human experts are engaged to check and assign a binary label to indicate correctness, mitigating noise cases that can arise when multiple feasible actions exist. We finally obtain a reliably curated set of 738 instances for this benchmark. Generally, our \name\ offers an end-to-end pathway to curate targeted data, train consistent and scalable critics, and evaluate them rigorously, with the goal of improving task success rates for computer-using agents in real environments.

We summarize contributions as follows:

(1) We propose a scalable data pipeline for synthesizing cross-platform GUI critic data, generating a dataset with $\sim$160k positive and $\sim$150k negative samples. It enables training critic models across diverse platforms.

(2) We introduce an elaborate training recipe that integrates supervised fine-tuning with consistency-preserving group relative policy optimization. Training with this strategy, our critic model, \ie, \name-7B, exhibits strong consistency on reasoning and judgment.

(3) We present OS-Critic Bench, a comprehensive benchmark designed for evaluating critic model performance across mobile, web, and desktop platforms. 

(4) The experiments demonstrate that our model sets new state-of-the-art results among open-source VLMs. 
When integrated with UI-TARS-1.5-7B, our critic model yields consistent performance gains across multiple online benchmarks (\eg, 29.2\% to 31.0\% on OSWorld).

\section{Related Work}
\label{sec:related_work}
\textbf{Computer-Using Agents.}
Building on the rapid progress of general-purpose Vision-Language Models~(VLMs)~\cite{bai2025qwen25vl,wang2025internvl35,hurst2024gpt,anthropic2025claude37,comanici2025gemini}, computer-using agents~(CUAs) have made significant advances in interacting with operating system environments, which enable VLMs serving as copilots to complete digital tasks~\cite{cheng-etal-2024-seeclick,sun2024genesis,zhang2025breaking,xie2025scalingcomputerusegroundinguser,cua2025,anthropic2024computeruse}.
Native agents~\cite{qin2025uitars,liu2025scalecua}  integrate planning and grounding within a single, end-to-end model.
These agents interact directly with the environment based on user instructions with raw visual observations~\cite{xu2024aguvis,wu2024atlas,wang2025opencuaopenfoundationscomputeruse,wang2025ui}.
As an alternative paradigm, modular agent frameworks~\cite{wu2024copilot,agashe2024agent,yang2025gta1,song2025coact,gonzalez2025unreasonable,xie2025scalingcomputerusegroundinguser} consist of several sub-agents for planning, reflection, and memory, respectively, which exhibit promising computer-using performance. 
Despite that, current CUAs~\cite{ye2025mobile,liu2025scalecua,wang2025opencuaopenfoundationscomputeruse,qin2025uitars,wang2025ui,nguyen2025gui} still frequently generate infeasible or inaccurate actions that lead to task failure, underscoring an urgent need for a step-level critic model tailored to CUAs.

\noindent\textbf{Critic Models for CUAs.}
Critic models play an important role in CUAs across data collection, model training, and online task completion~\cite{wang2025opencuaopenfoundationscomputeruse,ye2025mobile,gu2025ui,wang2025ui,lin2025cuarewardbench,men2025agent,sun2025seagent}. Existing critics for CUAs fall into two families.
Broadly, existing work instantiates critics in two forms: general-purpose VLMs used as in-context critics, and instruction-tuned VLMs specialized for computer-using. 
%
On the one hand, OS-Genesis~\cite{sun2024genesis} and NNetnav~\cite{murty2024nnetscape} prompt off-the-shelf VLMs to score raw trajectories, yielding higher-quality training data, and some systems~\cite{yang2025zerogui,guiactor} adopt general-purpose VLMs as external reflection modules during online interaction. However, such in-context critics based on general-purpose VLMs exhibit strong acceptance bias and can produce unstable judgments~\cite{men2025agent,lin2025cuarewardbench}. 
On the other hand, a line of work trains VLMs as reward models tailored for CUAs~\cite{sun2025seagent,wanyan2025look,xiao2025ui,qin2025uitars,sun2025sentinel}. For instance, UI-TARS-2~\cite{wang2025ui} adopt the model itself as an outcome reward model~(ORM) to validate web tasks in RL, UI-Critic-R1~\cite{wanyan2025look} and UI-Genie~\cite{xiao2025ui} further provide step-wise feedback at inference time, but are trained solely on mobile data, limiting cross-platform generalization. In contrast, we train a unified critic model that delivers step-level, reasoning-grounded judgments across mobile, web, and desktop, to support the full CUA lifecycle.

\noindent
\textbf{Computer-Using Benchmarks.}
A variety of benchmarks have emerged to assess computer-using abilities on several aspects, including GUI content understanding~\cite{wang2025mmbenchgui}, GUI grounding~\citep{cheng-etal-2024-seeclick,wu2024atlas,li2025screenspotpro,wang2025mmbenchgui,xie2025scalingcomputerusegroundinguser}, and GUI task completion~\citep{rawles2023androidinthewild,guicourse,androidcontrol,deng2023mind2web,kapoor2024omniact,lu2024gui_odyssey,zhou2023webarena,xie2024osworld,bonatti2024windows,rawles2024androidworld,wang2025mmbenchgui,sun2025scienceboard}.
In addition, some benchmarks focusing on critic evaluation for computer use agents are proposed. GUI-Critic-Test~\cite{wanyan2025look} is built from open-source data and is thus vulnerable to data leakage, while AgentRewardBench~\cite{men2025agent} and CUARewardBench~\cite{cua2025} are manually constructed but restricted to the web and desktop domains, respectively. 
In contrast, our OS-Critic Bench is curated from CUAs' inference trajectories with human annotation, covering the mobile, web, and desktop domains. We believe it can provide a comprehensive and precise evaluation of models' critic capabilities for computer-using agents.
\section{OS-Oracle Data Pipeline}
\label{sec:data_pipeline}

\begin{figure*}[ht]
  \centering
  \includegraphics[width=0.9\linewidth]{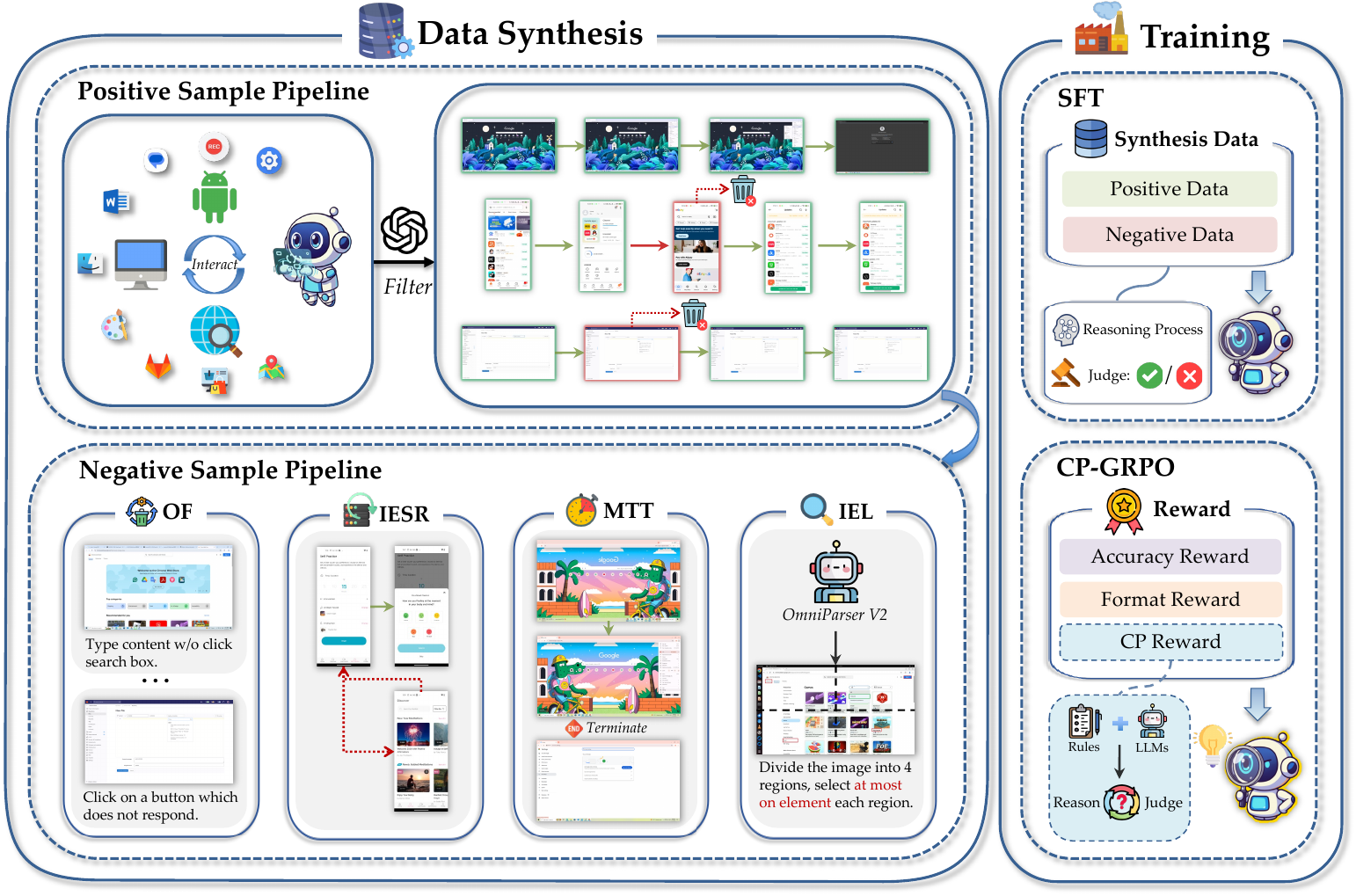}
\caption{The framework of OS-Oracle. (Left) Data Synthesis. We first use GPT-based filtering to extract step-wise positive samples from raw trajectories, and then construct four types of negative samples from the positive ones: Operation Failure (OF), Inefficient Error State Recovery (IESR), Mistimed Task Terminatio (MTT), and Inaccurate Element Localization (IUEL). (Right) Two-stage training: supervised fine-tuning (SFT) on synthesized data followed by CP-GRPO training with accuracy, format, and consistency-preserving rewards.}
  \label{fig:framework}
\end{figure*}

\subsection{Background}
\label{sec:background}
\textbf{CUAs Formulation.} 
The process of GUI automation task can be formalized as a Partially Observable Markov Decision Process (POMDP) $\left \langle g,\mathcal{S},\mathcal{A},\mathcal{O},\mathcal{T}\right \rangle$. Here, $g$ denotes the task goal, $\mathcal{S}$ the state space, $\mathcal{A}$ the action space, $\mathcal{O}$ the observation space, and $\mathcal{T}: \mathcal{S} \times \mathcal{A} \to \mathcal{S}$ the state transition function. At each time step $t$, the agent executes a decision based on policy $\pi$, which integrates the task goal $g$, memory $m_t=\left \{ o_j,a_j,o_{j+1},a_{j+1},\dots,o_{t-1}, a_{t-1} \right \},0\le j < t$, and the current observation $o_t$. The agent's trajectory, $\tau = \left [ s_0, a_0, s_1, a_1, \dots, s_t \right ]$, emerges from the policy and state transitions, as formulated by:
\begin{equation}
p_{\pi}(\tau) = p\left(s_{0}\right) \prod_{t=0}^{T} \pi\left(a_{t} \mid g, s_{t}, m_{t}\right) \mathcal{T}\left(s_{t+1} \mid s_{t}, a_{t}\right).
\end{equation}
\textbf{Critic Models.} Approaches for critic models generally fall into two categories: (i) Pointwise scoring, where the model evaluates each candidate response independently; and (ii) Pairwise ranking, which identifies the better response from a given pair of candidates. We adopted the pointwise scoring scheme because a pairwise critic model is ill-suited for GUI tasks, which is forced to determine a preference between two actions, even if both are functionally ineffective. The pointwise scoring, in contrast, is trained to evaluate each action independently against an objective standard of correctness, providing a clear and interpretable score of its validity for task completion. 
A pointwise critic model $\mathcal{M}_{\mathrm{critic}}$ for CUAs is formulated to assess the quality or correctness of an agent's action $a_t \in \mathcal{A}$ at a given time step $t$:
\begin{equation}
r_t, j_t = \mathcal{M}_{\mathrm{critic}}(g, m_t, o_t, a_t),
\end{equation}
where $r_t$ represents the reason for action assessment and $j_t$ represents the judgment, which can be a continuous value in the range $[0, 1]$, or a binary value of 0 or 1. Our primary objective is to determine step-level validity. Therefore, we define the critic's output as a binary value \(j_i\in\{\text{Yes},\text{No}\}\), where the former signifies a correct action and \text{No} signifies an incorrect one. Figure \ref{fig:teaser} illustrates the overall workflow of OS-Oracle-7B, showing how the critic model analyzes task context, history, proposed action, and current GUI state to determine correctness.

\subsection{Data Construction}
\label{sec:pos-gen}
To successfully transform VLMs to the binary pointwise critic models, a central bottleneck is the extreme scarcity of requisite training data, particularly high-quality negative samples. Publicly available datasets for GUI planning are already scarce, and the vast majority consist of expert demonstrations which are composed almost entirely of correct operations, leading to an inherent lack of negative samples. Moreover, distinguishing between correct and erroneous operations within these trajectories generated by CUAs is exceptionally challenging. Even when a CUA successfully completes a task, its execution path may contain numerous redundant, inefficient, or unnecessary steps. Conversely, for trajectories where the CUA fails, precisely identifying the specific erroneous step responsible for the failure is notoriously difficult in practice. Consequently, a straightforward strategy based solely on the final task outcome to determine correctness of steps is highly unreliable. Compounding this issue, the lack of robust GUI planning knowledge in current open-source VLMs  makes correctness judgement diffcult and futher exacerates negative sample collection. These factors collectively impede the development of GUI critic models capable of effectively identifying operational errors. 

To address the aforementioned bottleneck, we have meticulously designed an automated negative sample construction pipeline as shown in Figure~\ref{fig:framework}, which bypass the difficult requirement of explicitly judging the correctness of each step within a trajectory.
We analyze the key error patterns of CUAs and simulate these typical failure scenarios to automatically construct high-fidelity negative samples from existing trajectory datasets. Specifically, our pipeline targets several critical deficiencies observed in CUAs, and constructs corresponding erroneous interaction contexts based on existing trajectory datasets to help the critic effectively distinguish these errors.
%

\textbf{Operation Failure~(OF).} 
It addresses the lack of perception for subtle state changes by simulating three key error scenarios: (1) synthetically inserting `type' actions before their corresponding `click' actions, to simulate the failure to perceive the subtle visual activation of an input field; (2) artificially repeating eligible click operations, to simulate the failure to register minimal, localized UI changes post-click; and (3) appending redundant `scroll up' or `scroll down' actions, to simulate the failure to perceive the absence of visual change when at a scroll boundary.

\textbf{Inefficient Error State Recovery~(IESR).} 
The pipeline targets the lack of recovery capabilities from error states. Agents often fail to execute a `back' operation or other corrective measures when encountering an unexpected UI state. 
We address this using a state-injection method. For a given correct step, we first use its observation to retrieve trajectories with highly similar observations and then randomly select one of these trajectories to extract its subsequent step injected into the given step as a synthetic error next state, where we posit that all operations except for the 'back' action should be considered negative samples. 
%

\textbf{Mistimed Task Termination~(MTT).} 
We address the insufficient understanding of task termination states, where agents may not recognize when a task is successfully completed. We construct negatives by either appending redundant operations to a complete, successful trajectory or by prematurely truncating an unfinished trajectory and appending a terminate action. 

\textbf{Inaccurate Element Localization~(IEL).} 
CUAs may choose the correct action type but apply it to the wrong UI element. To simulate this failure mode, we first use OmniParser V2~\cite{yu2025omniparser} to detect all interactive elements. When metadata such as the AndroidControl accessibility tree is available, we extract element information from it and compute the Intersection over Union (IoU) between metadata-defined and detected elements, retaining detected elements with IoU \(> 0.7\). The remaining detected UI elements are then partitioned into a 2x2 grid over the image, and we sample at most one element per cell to obtain diverse error candidates.

Through the four error patterns corresponding data construction, we obtained erroneous operations as negative samples. With operations from existing trajectories as positive samples, we consequently derived a prompt set suitable for training critic models.
\subsection{Data Annotation}

Given an initial dataset comprising agent interaction contexts and their corresponding judgements, 
we employ GPT-4o~\cite{hurst2024gpt} to generate explicit reasons for these judgements. 
This crucial step, termed as rationale generation, effectively bootstraps the originally sparse labels into an information-rich set of explanatory rationales. 
Training on this enriched data enables the model to internalize the evaluation process rather than merely mimic final outcomes, resulting in more robust alignment.

Specifically, since the negative samples are constructed for specific failure modes, we provide GPT-4o with the task history, incorrect actions, and the corresponding error category and cause. This additional signal helps GPT-4o generate more precise rationales and better align its criticism with the task context, thereby improving the capacity of the critic models. Prompt templates are given in the Appendix. We discard samples where GPT-4o's judgment of operational correctness conflicts with the ground-truth labels to ensure annotation quality. Using this pipeline, we construct the OS-Oracle training set with 160k positive and 150k negative samples.

\section{OS-Oracle Training}
Based on our critic datasets, we propose a two-stage training strategy to effectively train the OS-Oracle critic model, combining supervised fine-tuning and a novel reinforcement learning objective.
\subsection{Supervised Fine-tuning}

The training pipeline begins with Supervised Fine-Tuning (SFT), which serves as the foundational stage for initializing the model with task-specific knowledge. 
Starting from a VLM, 
we fine-tune the model parameters $\theta$ on the OS-Oracle training dataset $\mathcal{D} = \{(x_i, r_i, j_i)\}$, where each input $x_i = (g, m_t, o_t, a_t)$ represents a GUI context. The objective is to minimize the cross-entropy loss for both the reason generation and judgment prediction:
\begin{equation}
\mathcal{L}_{\text{SFT}} = -\mathbb{E}_{(x, r, j) \sim \mathcal{D}} \left[ \log P_{\theta}(r \mid x) + \log P_{\theta}(j \mid x, r) \right].
\end{equation}
This stage ensures that the model acquires a robust initial ability to generate plausible reasoning traces and correct judgments before advancing to reinforcement learning.

\subsection{Consistency-Preserving GRPO}
\label{sec:cp-grpo}

To further enhance the discriminative capability and alignment of the model, we proceed to the second stage where this stage refines the SFT model $\pi_{\theta}^{\text{SFT}}$ into an improved policy $\pi_{\theta}$ through reinforcement learning. We build upon the Group Relative Policy Optimization (GRPO) framework, which promotes relative performance improvements within semantically similar query groups. GRPO employs a group-based approach: for a group of $N$ semantically similar queries $\mathcal{G} = \{x_1, x_2, \ldots, x_N\}$, we sample outputs $\{\hat{r}_i, \hat{j}_i\} \sim \pi_{\theta}(\cdot \mid x_i)$ and compute rewards that facilitate relative comparisons within the group. The optimization objective is to maximize the expected reward:
\begin{equation}
\mathcal{J}_{\text{RL}}(\theta) = \mathbb{E}_{x \sim \mathcal{D}} \left[ \mathbb{E}_{\hat{r}, \hat{j} \sim \pi_{\theta}(\cdot \mid x)} \left[ R(x, \hat{r_i}, \hat{j_i}) \right]\right],
\end{equation}
where $R(x, \hat{r_i}, \hat{j_i})$ represents the reward function.
During GRPO training, we observed a notable inconsistency between generated reasons and final judgments: the model sometimes produces reasons that logically support a correct action, yet the final judgment incorrectly classifies the action as wrong. 

To address this issue, we introduce Consistency-Preserving GRPO (CP-GRPO) incorporating a consistency reward $R_{\text{consistency}}$ via a hybrid rule-based and model-based scheme, which explicitly encourages alignment between the reasoning content and the judgment outcome.

Specially, we first employ a rule-based method to determine the polarity of a predicted reason $\hat{r}_i$ based on a positive lexicon $\mathcal{L}^{+}=\{\text{relevant},\text{valid},\text{align with},...\}$ ,indicating the action is helpful for completing the task and a negative lexicon $\mathcal{L}^{-}=\{\text{incorrect}, \text{not aligned with},\text{error},...\}$ which represents the action cannot help complete the task. The reasons is extracted and segment it into semantically continuous units. Then, we respectively count the number of occurrences of these units from the positive and negative lexicons:
\begin{equation}
    c_i^{+}=\!\!\sum_{w\in \mathcal{T}(\hat{r}_i)}\!\!\mathbf{1}[w\in \mathcal{L}^{+}],\quad
c_i^{-}=\!\!\sum_{w\in \mathcal{T}(\hat{r}_i)}\!\!\mathbf{1}[w\in \mathcal{L}^{-}],
\end{equation}
where $\mathbf{1}$ is the indicator function and $\mathcal{T}$ represents split function. The resulting score is $s_i^{\mathrm{rule}}=c_i^{+}-c_i^{-}$, which determines the rule-based polarity $p_i^{\mathrm{rule}}$ based on
$p_i^{\mathrm{rule}} = \operatorname{sgn}(s_i^{\mathrm{rule}})$, where $\operatorname{sgn}(0) = \varnothing$.
If $p_i^{\mathrm{rule}}=\varnothing$, we fall back to a model-based approach, where we feed Qwen3-8B with the reason to directly generate a proxy judgment and polarity $p_i^{\mathrm{model}}$ is determined by comparing the logits:
\begin{equation}
    p_i^{\mathrm{model}}=
\begin{cases}
+1,& \ell_i^{\text{Yes}}>\ell_i^{\text{No}},\\
-1,& \ell_i^{\text{Yes}}\le \ell_i^{\text{No}},
\end{cases}
\end{equation}
where $\ell_i^{\text{Yes}}$ and $\ell_i^{\text{No}}$ denote the logits for Yes and No, respectively. The final polarity for the reason is calculated as: 
\begin{equation}
    \tilde{p}_i =
\begin{cases}
p_i^{\mathrm{rule}},  & \text{if } p_i^{\mathrm{rule}} \neq \varnothing, \\
p_i^{\mathrm{model}}, & \text{if } p_i^{\mathrm{rule}} = \varnothing.
\end{cases}
\end{equation}
We define the consistency reward $R_{\text{consistency}}$ as 1 if $\tilde{p}_i$ matches judgment $j_i$, and 0 otherwise, as follows:
\begin{equation}
    R_{\text{consistency}}(x_i,\hat{r}_i,\hat{j}_i)=\mathbf{1}\!\left[\tilde{p}_i=\nu(\hat{j}_i)\right],
    \label{eq:cons}
\end{equation}
where $\nu(\text{Yes}){=}{+}1$ and $\nu(\text{No}){=}{-}1$.
The overall CP-GRPO reward can be formulated:
\begin{equation}
R(x_i, \hat{r}_i, \hat{j}_i) = \alpha R_{\text{acc}} + \beta R_{\text{format}} + \gamma R_{\text{consistency}},
\label{eq:cp-grpo-reward}
\end{equation}
where the accuracy reward $R_{\text{acc}}$ is defined as $\mathbf{1}[\hat{j}_i = j_i]$ and format reward $R_{\text{format}}$ is a binary score indicating whether the generated reason $\hat{j}_i$ conforms to the required structure.

\section{Experiments}
\label{sec:experiments}

\begin{table*}[t]
  \caption{Performance comparison on OS-Critic Bench. The table presents proprietary models, open-source models, and \name. Higher values are better for all metrics (Accuracy and F1-Score) across Mobile, Desktop, Web, and Overall settings. The \textbf{best} and \underline{second-best} results are indicated in bold and underlined, respectively.}
  \label{tab:main_results}
  \begin{tabular*}{\textwidth}{@{}l@{\extracolsep{\fill}}cccccccc@{}}
    \toprule
    \multirow{2}{*}{\textbf{Model}}& \multicolumn{2}{c}{\textbf{Mobile}}  & \multicolumn{2}{c}{\textbf{Desktop}} & \multicolumn{2}{c}{\textbf{Web}} & \multicolumn{2}{c}{\textbf{Overall}} \\
    
    \cmidrule(lr){2-3} \cmidrule(lr){4-5} \cmidrule(lr){6-7} \cmidrule(lr){8-9}
    & \makecell[c]{Accuracy} & \makecell[c]{F1-Score}  & \makecell[c]{Accuracy} & \makecell[c]{F1-Score} & \makecell[c]{Accuracy} & \makecell[c]{F1-Score} & \makecell[c]{Accuracy} & \makecell[c]{F1-Score}  \\
    
    \midrule
    \multicolumn{9}{c}{\textbf{Proprietary Models}} \\
    \midrule
     GPT-4o~\cite{hurst2024gpt} & 62.79 & 66.53 & 63.82 & 70.90 & 58.78 & \underline{65.92} & 62.20 & 67.37 \\
     GPT-5~\cite{gpt_5} & \textbf{69.18} & \underline{69.53} & \underline{73.68} & \underline{74.36} & 59.46 & 55.22 & \textbf{68.16} & \underline{67.94} \\
     Claude-4.5-Sonnet~\cite{claude_45} & 65.30 & 63.46 & \textbf{74.34} & \textbf{77.46} & \underline{64.19} & 64.90 & 66.94 & 67.03 \\
     Gemini-2.5-Pro~\cite{comanici2025gemini} & \underline{68.26} & \textbf{70.11} & 67.11 & 70.24 & \textbf{66.22} & \textbf{70.24} & \underline{67.62} & \textbf{70.16}  \\    
    \midrule
    \multicolumn{9}{c}{\textbf{Open Source Models}} \\
    \midrule
     Qwen2.5-VL-7B~\cite{bai2025qwen25vl} & 58.45 & 65.79 & 61.18 & 69.43 & 54.73 & 64.17 & 58.27 & 66.23 \\                 
     Qwen3-VL-8B~\cite{Qwen3-VL} & 61.64 & 62.33 & 59.87 & 63.91 & 60.14 & 60.40 & 60.98 & 62.30 \\
     UI-TARS-1.5-7B~\cite{qin2025uitars} & 56.16 & 66.43 & 58.55 & 69.57 & 60.14 & 70.05 & 57.45 & 67.83 \\
     ScaleCUA-7B~\cite{liu2025scalecua} & 50.23 & 59.93 & 44.74 & 58.82 & 34.46 & 49.21 & 45.93 & 57.51 \\
     ScaleCUA-32B~\cite{liu2025scalecua} & 51.60 & 63.32  & 49.34  & 64.84  & 50.68 & 65.40  & 50.95 & 64.09 \\
     SE-WSM-7B~\cite{sun2025seagent} & 50.91 & 16.99 & 50.66 & 13.79 & 41.22 & 18.69 & 48.92 & 16.78 \\
     GUI-Critic-R1~\cite{wanyan2025look} & 59.13 & 66.54 & 59.21 & 70.19 & \underline{60.81} & \textbf{72.90} & 59.49 & 68.76 \\
     \midrule
     \name-7B-SFT & \underline{63.47} & \underline{71.22} & \underline{63.16} & \textbf{72.00} & \textbf{62.16} & \underline{71.72} & \underline{63.14} & \underline{71.49} \\
     \name-7B & \textbf{70.78} & \textbf{74.30} & \textbf{65.79} & \underline{71.11} & \textbf{62.16} & 70.53 & \textbf{68.02} & \textbf{72.81} \\
    
    \bottomrule
  \end{tabular*}
\end{table*}

\subsection{OS-Critic Bench}
To evaluate the versatility and generalization of critic models. We introduce \textbf{OS-Critic Bench}, the first cross-platform benchmark underwent rigorous quality control.

\textbf{Data Source.} To ensure comprehensive cross-platform evaluation, the benchmark construction utilizes data from three distinct domains. The Mobile component is sourced from the AndroidControl~\cite{androidcontrol} test set and the GUIOdyssey~\cite{lu2024gui_odyssey} test set. For the Web platform, we incorporate data from the guiact~\cite{guicourse} test set, supplemented by a non-overlapping subset of ScaleCUA-Web~\cite{liu2025scalecua} that is reserved exclusively for benchmarking. Finally, Desktop scenarios are derived from AgentNet-Bench~\cite{wang2025opencuaopenfoundationscomputeruse}.

\textbf{Operation Sampling.} We utilize these task prompts and their corresponding screenshots to sample candidate operations using the Qwen2.5-VL-7B-Instruct model, thus yielding a large amount of data available for evaluation by the critic model. 

\textbf{Quality Control.} Crucially, due to the inherent non-unique nature of solutions in complex GUI-based tasks, we departed from traditional rule-based methods that rely strictly on comparing generated actions against existing ground truth actions for judgment. Instead, we introduced  human experts was tasked with providing the final and definitive judgment for the correctness of the sampled actions, even if the action deviated from the original trajectory ground truth but still constituted a valid step towards the task goal. This meticulous, human-centric validation process ensured the high quality and fidelity of the resulting dataset, yielding a substantial volume of reliable data available for robust evaluation by the critic model.

Through this process, we collected 738 samples for evaluation. Each instance in the dataset comprises a task goal, memory, a screenshot, an action under review, a concise prompt serving as a critic, and a human-annotated binary label indicating whether the action advances the task completion.

\subsection{Setup}
\textbf{Training.}
We adopt Qwen2.5-VL-7B-Instruct as the backbone and train it in two consecutive phases. During the SFT phase, we train on the entire OS-Oracle dataset with 1 epoch, resulting in \name-7B-SFT. Subsequently, in the CP-GRPO phase, we further optimize \name-7B-SFT on the same dataset for 3 epochs to obtain \name-7B, where the KL regularization coefficient is set to $3\times10^{-5}$ with reward weights $\alpha=0.9$, $\beta=0.05$, and $\gamma=0.05$ in Eq.~(\ref{eq:cp-grpo-reward}). The rollout batch size and group size are 512 and 16, respectively.

\noindent\textbf{Evaluation.} We comprehensively evaluate the capabilities of \name-7B across three aspects: (1) \textbf{Offline Evaluation}: Tests its judgment capability to compare agent's prediction with predefined actions. (2) \textbf{Dynamic Evaluation}: Assesses its online assistance capability by measuring its ability to guide an agent toward task completion, showing its practical utility. (3) \textbf{Data Quality Control}: Probes its potential for automated data quality control on GUI datasets. 

\subsection{Main results}
\textbf{Offline Evaluation.}
We evaluate a diverse set of general-purpose VLMs and CUAs on the OS-Critic Bench shown in Table~\ref{tab:main_results}. Proprietary VLMs, particularly GPT-5 and Claude-4.5-Sonnet, produce impressive results on Desktop while there is still potential for enhancement with regards to Web and Mobile. 
Current open-source VLMs, such as Qwen2.5-VL-7B and Qwen3-VL-8B, exhibit certain critic capabilities thanks to their inherent reasoning ability, enabling them to assess GUI actions to some extent even without task-specific training.
In contrast, existing CUAs such as ScaleCUA-7B and ScaleCUA-32B, while proficient in performing GUI operational tasks, are inherently less capable of evaluating the validity of actions, leading to noticeably lower critic performance compared to general-purpose VLMs.
SE-WSM-7B, a world state model for CUAs, exhibits poor performance and generalization because it is trained on a dataset with limited diversity. Similarly, GUI-Critic-R1, although specifically designed as a critic model for GUI scenarios, shows only marginal improvement over its base model Qwen2.5-VL-7B due to the constrained domain coverage and limited scale of its training data.
Our \name-7B achieves the highest accuracy among all open-source models and delivers the state-of-the-art performance on Mobile, which emphasize the importance of extensive training for GUI critic models and confirm that our large-scale dataset generated by a scalable data pipeline can achieve superior performance and generalization. 

\begin{figure}[!ht]
  \centering
  \includegraphics[width=0.46\textwidth]{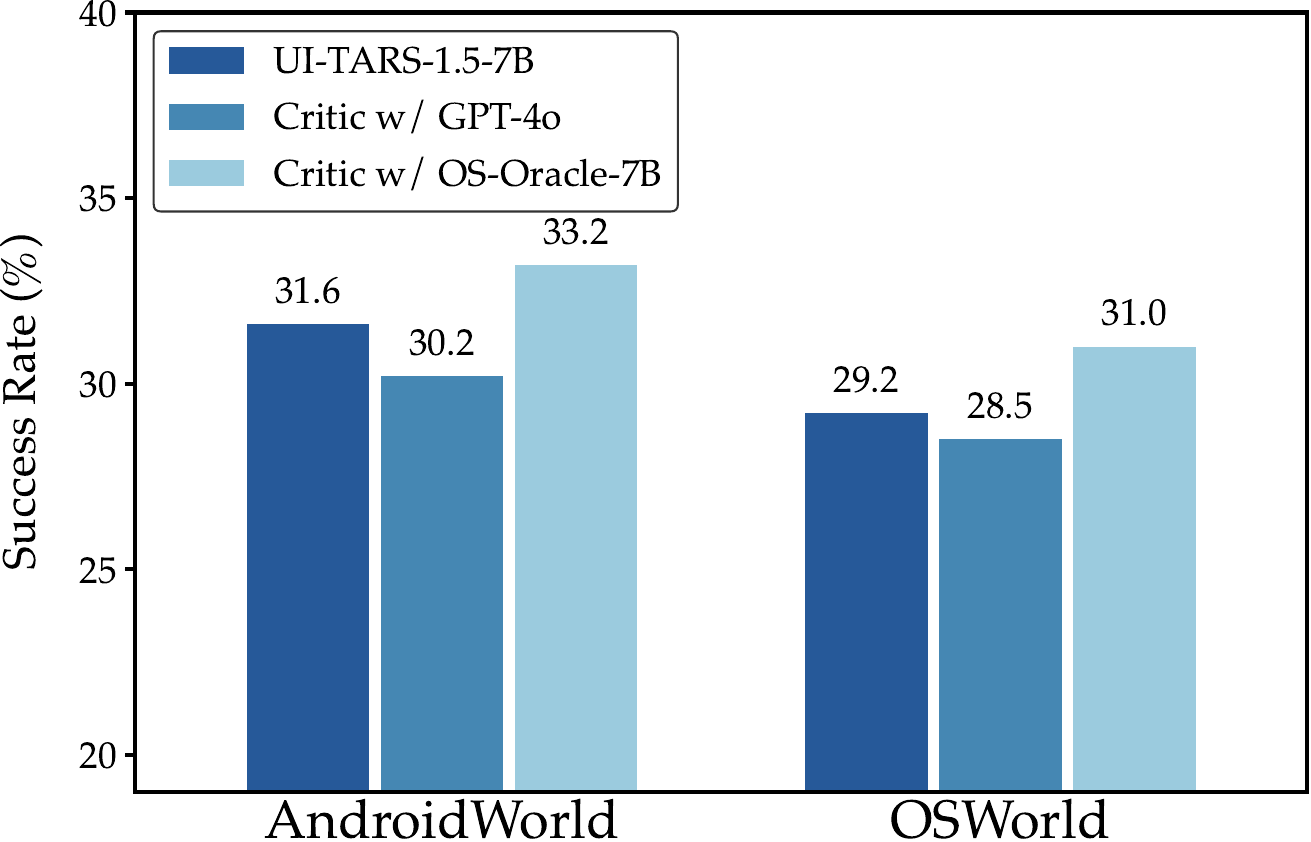}
  \caption{The performance of UI-TARS-1.5-7B with critic model on AndroidWorld and OSWorld.}
  \label{fig:critic_agent_results}
\end{figure}

\noindent\textbf{Dynamic Evaluation.} 
We compared GPT-4o and \name-7B as critic models to guide UI-TARS-1.5-7B in completing tasks within AndroidWorld and OSWorld. Specifically, if the critic model deemed an action incorrect, the CUAs were required to regenerate the action with a maximum of three attempts.
The evaluation results are shown in Fig.~\ref{fig:critic_agent_results}, which demonstrate that across both platforms, the performance of the base agent is enhanced by \name-7B, while GPT-4o tends to degrade the performance of CUAs due to a lack of relevant knowledge, resulting in inaccurate evaluations of operational elements and consequently leading the model into erroneous states.

\noindent\textbf{Data Quality Control.} We assessed the potential of the critic model to establish a data flywheel, which enables the collection of high-quality GUI trajectory data. 
Specifically, we initially employed GPT-4o to generate diverse tasks. These tasks served as instructions for two open-source CUA models (ScaleCUA-7B and UI-TARS-1.5-7B) to produce trajectories, yielding a total of 10K mobile operational samples.
\begin{table}[h]
  \centering
  \caption{Comparison of SFT performance on all rollout trajectories vs trajectories filtered by \name-7B.}
  \label{tab:rft_results}
  \begin{tabular}{lc}
    \toprule
    \textbf{Model} & \textbf{AndroidWorld} \\
    \midrule
    Qwen2.5-VL-7B & 10.34 \\
    \quad + SFT & 12.07 \\
    \quad + SFT w/ \name-7B & \textbf{15.52} \\
    \bottomrule
  \end{tabular}
\end{table}

\begin{table}[h]
  \centering
  \caption{Effect of Synthetic vs.\ GPT-Annotated Negative Samples.}
  \label{tab:ablation_neg_data}
  \begin{tabular}{lcc}
    \toprule
    \multirow{2}{*}{\textbf{Model}} & \multicolumn{2}{c}{\textbf{Overall}} \\ 
    \cmidrule(lr){2-3}
    & \makecell[c]{Acc} & \makecell[c]{F1} \\
    \midrule
    Qwen2.5-VL-7B & 58.27 & 66.23 \\
    \quad + w/ GPT-Annotated Negatives & 55.42 & 67.33 \\
    \quad + w/ \name\ Synthetic Negatives & \textbf{60.03} & \textbf{69.37} \\
    \bottomrule
  \end{tabular}
\end{table} 
These samples are then used to train Qwen2.5-VL-7B-Instruct under two settings: (1) standard SFT on the entire 10K samples; (2) critic-guided SFT using \name-7B, which utilizes \name-7B to judge the correctness of actions within each sample, and then performs SFT on the curated subset of trajectories assessed by the critic as being valuable for task progression.
To balance total training exposure, we train the baseline for 1 epoch on the full dataset and the critic-guided model for 2 epochs on the filtered subset. 
As shown in Table~\ref{tab:rft_results}, the critic-guided SFT improves the downstream AndroidWorld performance by 3.45\% over the SFT baseline. This demonstrates that \name-7B can reliably filter low-quality actions, making it suitable for improving the quality of collected trajectories.

\subsection{Ablation}
\textbf{The effect of synthetic negative data.} 
We evaluate the quality of our synthetic incorrect operational samples by comparing with negative samples identified by GPT-4o, which are commonly adopted in prior studies~\cite{wanyan2025look, sun2025seagent}. We randomly select 50K positive samples from the full dataset for both settings and pair them with either 50K GPT-annotated negatives or 50K \name\ synthetic negatives to ensure a fair comparison. 
The two SFT models, each trained on its corresponding dataset, are evaluated using the OS-Critic bench shown in Table~\ref{tab:ablation_neg_data}. 
The model trained with our synthetic negative samples clearly outperforms the one trained with GPT-annotated negatives. Moreover, the GPT-annotated negative group performs even worse than the baseline, suggesting that such naturally annotated negatives contain substantial noise rather than meaningful negative supervision.

\noindent\textbf{The effect of the amount of critic data.} To examine how the amount of critic data affects model performance, 
we perform SFT on four training sets of different scales, all based on the same initial model. Specifically, we used $\pm10\text{K}$, $\pm50\text{K}$, $\pm100\text{K}$ positive and negative samples, respectively, as well as the full training set, to train the critic models and reported their performance on the OS-Critic Bench.
\begin{figure}[ht]
  \centering
  \includegraphics[width=0.47\textwidth]{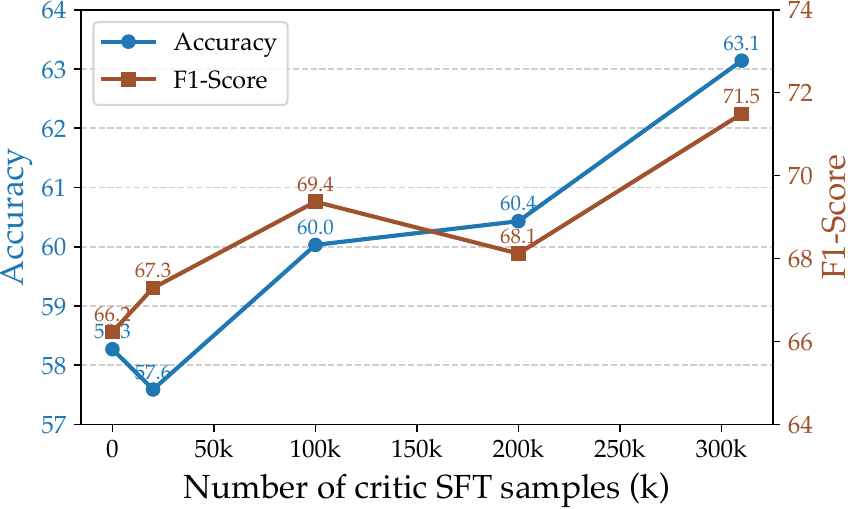}
  \caption{Effect of SFT data scaling on overall performance.}
  \label{fig:sft_data_scaling}
\end{figure}
\begin{table}[h]
  \centering
  \caption{Effect of consistency reward in CP-GRPO.}
  \label{tab:cons-reward}
  \begin{tabular}{lcc}
    \toprule
    \multirow{2}{*}{\textbf{Method}} & \multicolumn{2}{c}{\textbf{Overall}} \\ 
    \cmidrule(lr){2-3}
    & \makecell[c]{Acc} & \makecell[c]{Consistency} \\
    \midrule
    SFT + GRPO & 66.12 & 90.51 \\
    SFT + CP-GRPO & 65.31 & 99.59 \\
    \bottomrule
  \end{tabular}
\end{table}

\noindent\textbf{The effect of consistency reward.}
To assess the impact of incorporating a consistency reward into the GRPO objective, we utilize \name-7B-SFT as the initial model and then perform both GRPO and CP-GRPO training on the full dataset to compare the accuracy and the consistency calculated based on Eq~(\ref{eq:cons}). As detailed in Table~\ref{tab:cons-reward}, both methods achieve comparable accuracy while the consistency of GRPO is approximately 10\% lower than that of CP-GRPO. This disparity indicates that standard GRPO tends to induce mismatches between reasoning and judgment during GUI critic model training, whereas our CP-GRPO effectively mitigates these inconsistencies.

\section{Conclusion}
In this work, we present \textbf{\name}, a comprehensive framework for GUI critic models.  
By introducing a scalable cross-platform data pipeline, we systematically synthesize both positive and negative samples that capture diverse GUI failure modes. 
Together with a two-stage training recipe combining supervised fine-tuning and consistency-preserving GRPO, our approach enables robust and generalizable critic learning across Mobile, Web, and Desktop environments. Extensive experiments demonstrate that our critic model not only achieves impressive performance on the OS-Critic Bench but also effectively enhances the reliability and task success of native GUI agents. We hope this work will inspire future research and development of more capable and reliable GUI critic models.

{
    \small
    \bibliographystyle{ieeenat_fullname}
    \bibliography{main}
}
\clearpage
\maketitlesupplementary
\setcounter{section}{0}

\section{OS-Oracle Dataset Details and Statistics}
\label{sec:supply-dataset}




Unlike prior GUI critic datasets that focus on a single platform or interaction modality, as shown in Table~\ref{tab:dataset_statistics}, OS-Oracle provides a unified large-scale corpus that spans \textbf{desktop}, \textbf{web}, and \textbf{mobile} environments. 
Existing datasets typically emphasize only mobile or only desktop tasks, which limits a critic model’s ability to generalize across heterogeneous GUI ecosystems. 
In contrast, OS-Oracle offers balanced multi-platform coverage and serves as a comprehensive foundation for training robust cross-platform critics.

\begin{table}[htp]
  \footnotesize
  \centering
  \caption{CUA critic model dataset statistics.}
  \renewcommand{\arraystretch}{1.2}
  \begin{tabular}{l c c c c}
    \toprule
    \multicolumn{1}{c}{\multirow{2}{*}{\textbf{Data source}}} &
    \multicolumn{3}{c}{\textbf{Platform}} &
    \multicolumn{1}{c}{\multirow{2}{*}{\textbf{Samples}}} \\
    \cmidrule(lr){2-4}
    & \textbf{Mobile} & \textbf{Web} & \textbf{Desktop} & \\
    \midrule
    SEAgent~\cite{sun2025seagent}          & \xmark & \xmark & \cmark & 4,245 \\
    GUI-Critic-Train~\cite{wanyan2025look}    & \cmark & \xmark & \xmark & 11k \\
    UI-Genie-RM-517k~\cite{xiao2025ui} & \cmark & \xmark & \xmark & 517k \\
    \midrule
    Ours        & \cmark & \cmark & \cmark & 310k  \\
    \bottomrule
  \end{tabular}
  \label{tab:dataset_statistics}
\end{table}

Table~\ref{tab:os-oracle-data-statistics} presents an overview of OS-Oracle training corpus, detailing the positive and negative samples contributed by each dataset.
OS-Oracle synthesizes data from seven open-source datasets, resulting in a total of \textbf{162,760} positive samples and \textbf{150,667} negative samples. Overall, approximately \textbf{58.32\%} of the data comes from mobile environments, \textbf{23.17\%} from desktop, and \textbf{18.51\%} from web tasks.

\begin{table}[h]
    \centering
    \small
    \setlength{\tabcolsep}{4pt}
    \caption{OS-Oracle datasets statistics overview.}
    \renewcommand{\arraystretch}{1.2}
    \begin{tabular}{@{}l|c|l|l}
        \toprule
        \textbf{Training dataset}  & \textbf{Platform}  & \textbf{\#Positive} &  \textbf{\#Negative}    \\ \midrule
        AndroidControl  & Mobile & 52,466 & 52,162  \\
        AMEX  & Mobile & 9,370  &  7,154 \\
        AITZ  & Mobile  &  11,485 & 9,107 \\
        GUI-Odyssey  & Mobile  &  20,198  & 20,834 \\
        Mind2Web  & Web  & 6,326 & 6,578 \\
        ScaleCUA-Web  & Web  &22,846  &  22,270   \\
        AgentNet  & Desktop  & 40,069 & 32,562 \\ 
        \midrule
        \multicolumn{2}{c|}{\cellcolor[gray]{0.9}\textbf{Total}} & \cellcolor[gray]{0.9}\textbf{162,760} & \cellcolor[gray]{0.9}\textbf{150,667} \\
        \bottomrule
    \end{tabular}

    \label{tab:os-oracle-data-statistics}
\end{table}


\section{OS-Critic Bench}
OS-Critic Bench is a curated cross-platform evaluation suite comprising \textbf{738} carefully designed test cases spanning three major GUI environments: 438 mobile, 152 desktop, and 148 web tasks. The benchmark is constructed from representative sources across all platforms: for Mobile, the AndroidControl and GUIOdyssey test sets; for Web, the guiact test set and a dedicated, non-overlapping subset of ScaleCUA-Web; and for Desktop, AgentNet-Bench.
For each test case, the model receives a screenshot, the task instruction, the step-by-step action history, and a candidate action. The model must determine whether the candidate action is appropriate for completing the task, and its final verification is compared against human-annotated ground truth.
OS-Critic Bench assesses the model’s robustness under heterogeneous UI layouts, interaction modalities, and error patterns.

\begin{figure}[htp]
  \centering
  \includegraphics[width=0.4\textwidth]{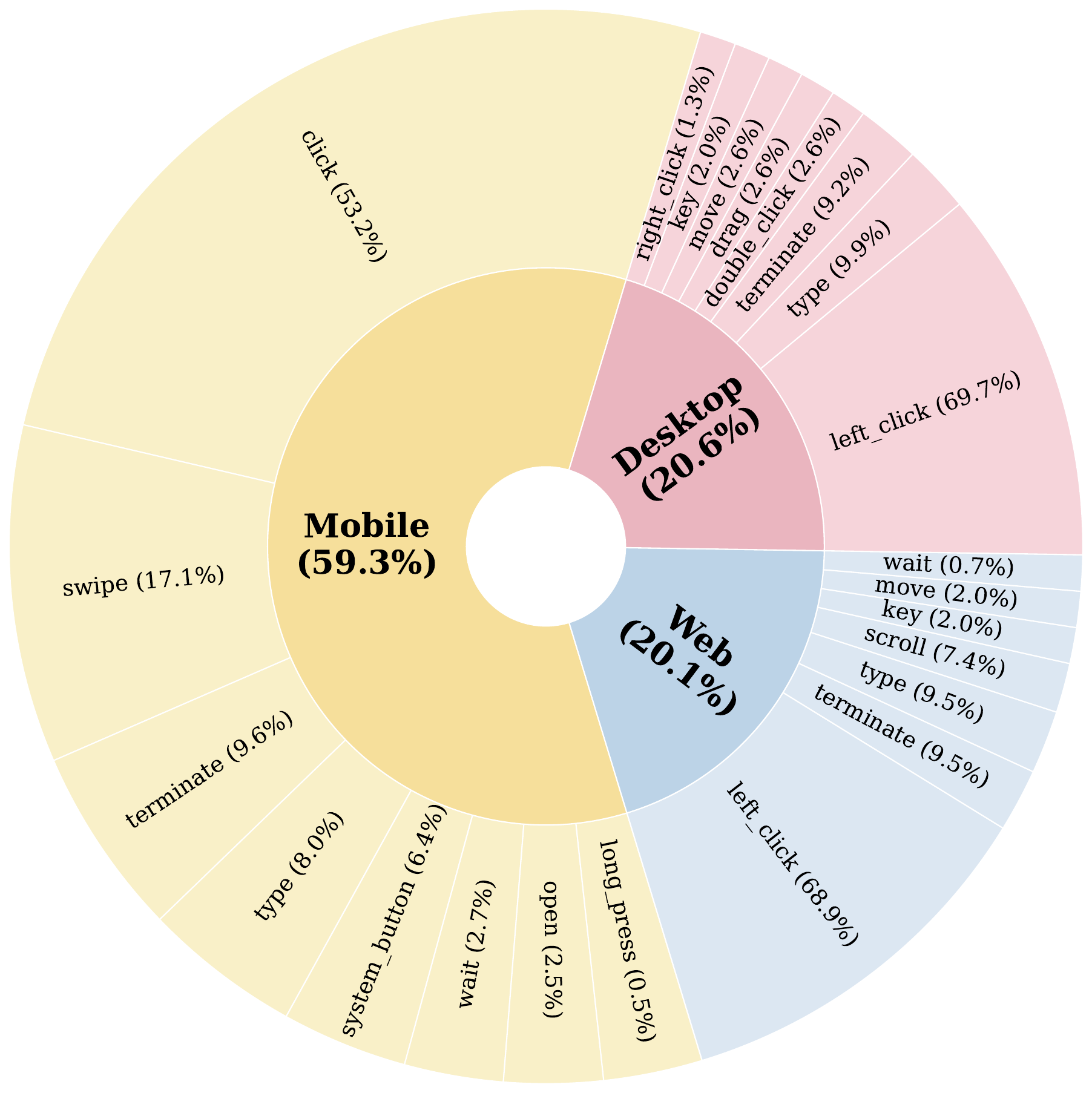}
  \caption{The action distribution of OS-Critic Bench.}
  \label{fig:os_critic_bench_action_distribution}
\end{figure}

As illustrated in Figure~\ref{fig:os_critic_bench_action_distribution}, OS-Critic Bench spans not only common GUI operations such as \texttt{click}, \texttt{type}, \texttt{scroll}, and \texttt{terminate}, but also platform-specific interaction patterns. 
For mobile environments, the benchmark includes actions such as \texttt{swipe}, \texttt{long\_press}, and \texttt{open}; 
for desktop scenarios, it incorporates operations like \texttt{right\_click}, \texttt{key}, \texttt{move}, and \texttt{drag}. 
To better align with everyday human–GUI behavior, we calibrate the action distribution on each platform so that the relative frequency of actions mirrors realistic usage patterns. We further enforce a balanced 1:1 ratio between positive and negative instances to mitigate distributional biases and discourage trivial hacks, making OS-Critic Bench a realistic and high-quality benchmark for critic evaluation.

\section{Unified Prompt}

We adopt a unified prompt format for both training and inference of \name-7B. 
Specifically, the mobile version of the prompt is provided in Table~\ref{tab:critic_prompt_mobile}, 
while the desktop and web versions are shown in Table~\ref{tab:critic_prompt_desktop}.

Across platforms, the prompt follows a consistent structure consisting of three main components:
\begin{enumerate}
    \item \textbf{Task framing and critic instructions}: clear guidance defining the critic’s role and expected output format.
    \item \textbf{Action space specification}: a detailed description of all permissible actions. 
    We align our definitions with the action spaces used in the Qwen2.5-VL mobile agent and computer-use agent implementations to ensure compatibility and coverage.
    \item \textbf{Task context}: including the current task instruction, the step-by-step action history, and the candidate action to be evaluated.
\end{enumerate}

This unified design ensures consistent behavior across platforms while capturing the unique interaction patterns of mobile, desktop, and web environments.

\section{Case Study}

\subsection{Pre-Critic in Online Benchmark}
To evaluate whether our critic model can provide reliable, step-level guidance in realistic execution settings, we conduct pre-critic experiments on both OSWorld and AndroidWorld online environments. In this setup, the base agent first proposes an action; the critic then evaluates whether the action is appropriate. If the critic judges the action as incorrect, the agent is prompted to regenerate an alternative action for up to three attempts before execution proceeds. This procedure reflects how critic models can be deployed to stabilize long-horizon GUI tasks and prevent irreversible errors.

\begin{figure*}[ht]
  \centering
  \includegraphics[width=0.8\linewidth]{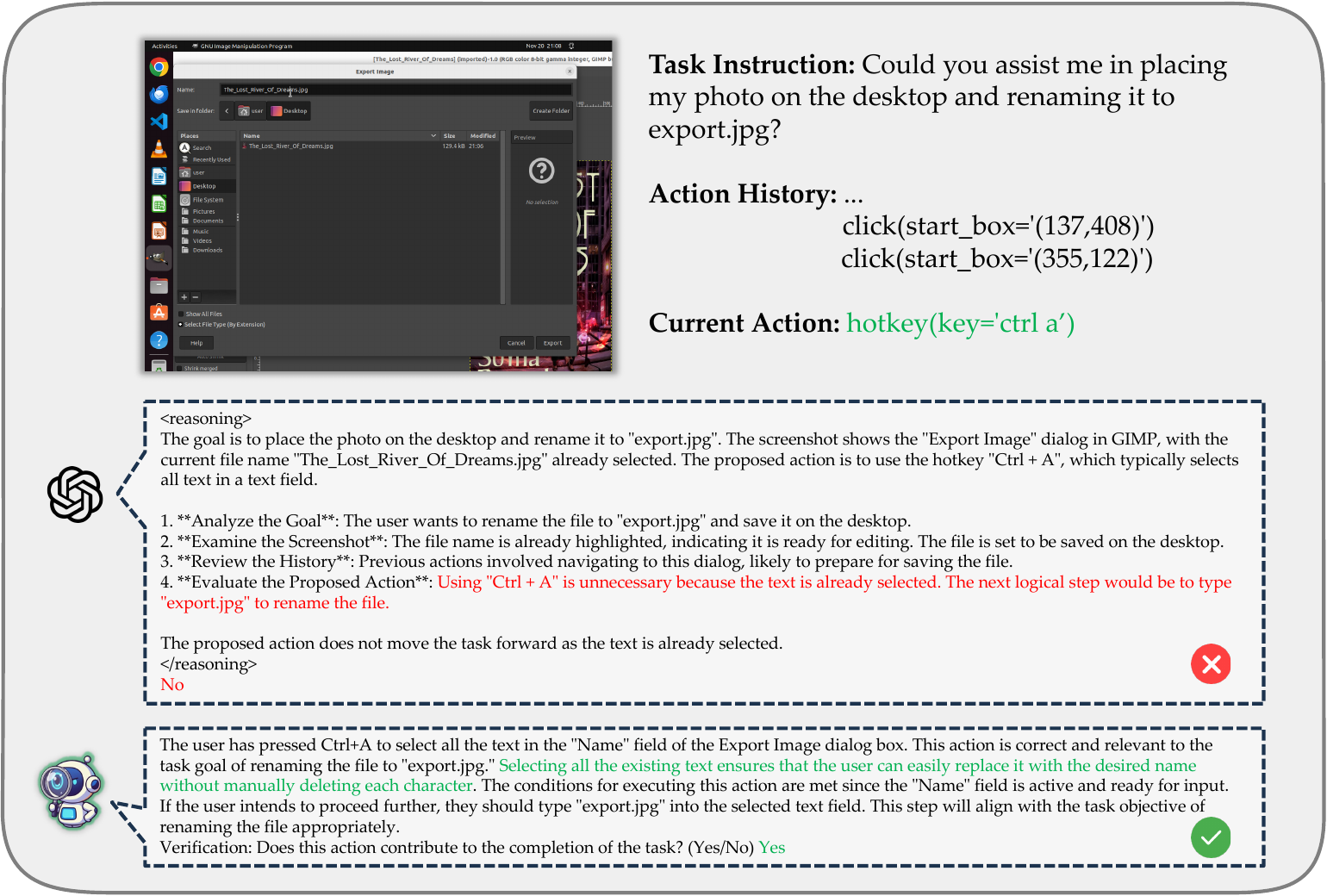}
\caption{An example on OSWorld comparing OS-Oracle-7B and GPT-4o as pre-critics. OS-Oracle-7B accurately assesses the screen state and the predicted action, whereas GPT-4o hallucinates and yields an incorrect critic judgment.}
  \label{fig:app-case-osworld}
\end{figure*}

\begin{figure*}[ht]
  \centering
  \includegraphics[width=0.8\linewidth]{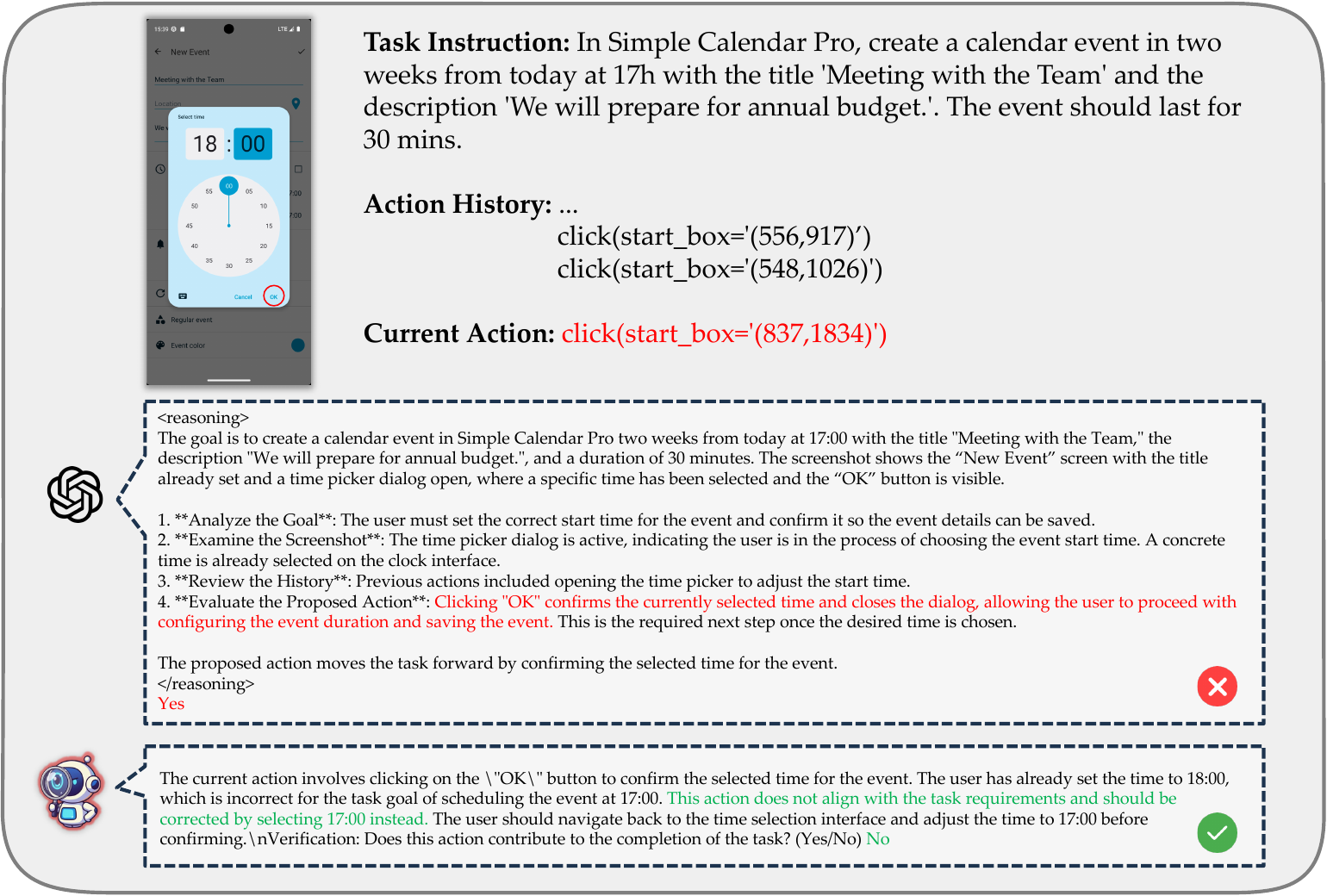}
\caption{An example on AndoridWorld comparing OS-Oracle-7B and GPT-4o as pre-critics. OS-Oracle-7B accurately assesses the screen state and the predicted action, whereas GPT-4o hallucinates and yields an incorrect critic judgment.}
  \label{fig:app-case-androidworld}
\end{figure*}

Fig.~\ref{fig:app-case-osworld} and Fig.~\ref{fig:app-case-androidworld} illustrate representative cases in OSWorld and AndroidWorld where GPT-4o provides hallucinated or incorrect critic judgments, while our OS-Oracle-7B offers accurate, state-grounded assessments.

\subsection{GRPO vs. CP-GRPO}
\begin{figure*}[ht]
  \centering
  \includegraphics[width=0.8\linewidth]{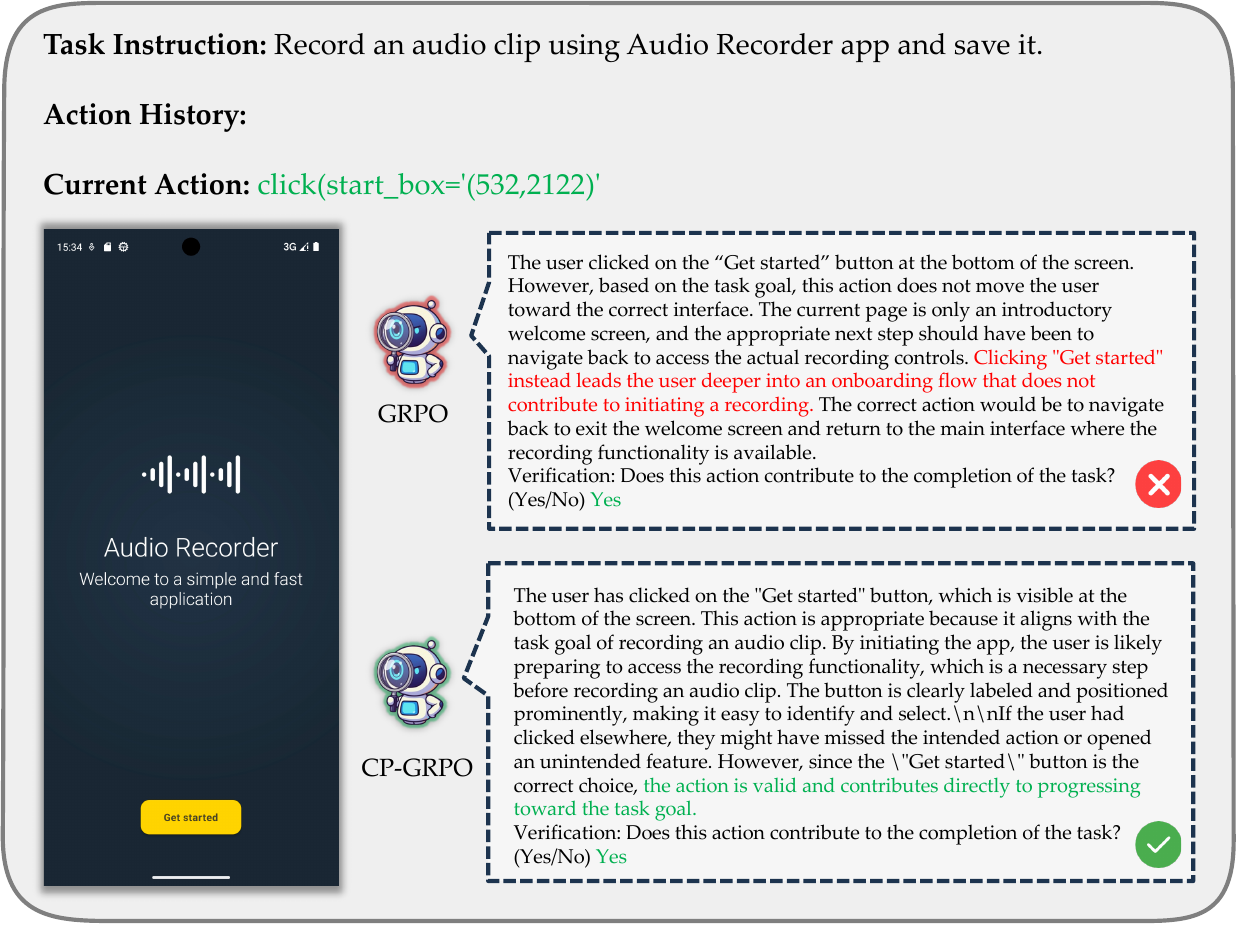}
\caption{Case Study: Comparing GRPO and CP-GRPO Critiques}
  \label{fig:case_study_grpo_vs_cpgrpo}
\end{figure*}

We conduct experiments between \name-7B and its GRPO-finetuned variant \name-7B-GRPO in AndroidWorld  Environment. As illustrated in Figure~\ref{fig:case_study_grpo_vs_cpgrpo}, \name-7B-GRPO occasionally produces critiques that are inconsistent with its final verification outcome—indicating a misalignment between intermediate reasoning and the final judgment. In contrast, \name-7B maintains stable and coherent behavior, delivering both correct critiques and consistent verification results. This example highlights the importance of consistency-preserving objectives such as CP-GRPO in stabilizing critic reasoning and ensuring reliable decision quality.


\onecolumn

\begin{longtable}{p{0.96\textwidth}}
\caption{Unified mobile critic prompt.}
\label{tab:critic_prompt_mobile} \\
\toprule
\textbf{Mobile Critic Prompt} \\
\midrule
\endfirsthead

\toprule
\textbf{Mobile Critic Prompt} \\
\midrule
\endhead

\bottomrule
\endfoot

You are an expert GUI task evaluator. Your role is to analyze the provided action in context, generate an insightful textual critique, and grade the action’s effectiveness.

Your evaluation must be based on the current screen image, the overall task instruction, and the history of previous actions.
At the end of your critique, you must provide a final grade in this exact format:
``Verification: Does this action contribute to the completion of the task? (Yes/No) X'', where X is either Yes or No.

\medskip
\textbf{Action Space Specification:}

\{

\hspace{0.3cm}``type'': ``function'',\\[2pt]

\hspace{0.3cm}``function'': \{\\

\hspace{0.6cm}``name\_for\_human'': ``mobile\_use'',\\
\hspace{0.6cm}``name'': ``mobile\_use'',\\[2pt]

\hspace{0.6cm}``description'': ``Use a touchscreen to interact with a mobile device, and take screenshots. This is an interface to a mobile device with touchscreen. You can perform actions like clicking, typing, swiping, etc. Some applications may take time to start or process actions, so you may need to wait and take successive screenshots to see the results of your actions. The screen’s resolution is \{SCREEN\_WIDTH\}x\{SCREEN\_HEIGHT\}. Make sure to click any buttons, links, icons, etc with the cursor tip in the center of the element. Don’t click boxes on their edges unless asked.'',\\[4pt]

\hspace{0.6cm}``parameters'': \{\\

\hspace{0.9cm}``properties'': \{\\

\hspace{1.2cm}``action'': \{\\
\hspace{1.5cm}``description'': ``The action to perform. The available actions are: ‘action=key’: Perform a key event on the mobile device. – This supports adb’s `keyevent` syntax. – Examples: ``volume\_up'', ``volume\_down'', ``power'', ``camera'', ``clear''. ‘action=click’: Click the point on the screen with coordinate (x, y). ‘action=long\_press’: Press the point on the screen with coordinate (x, y) for specified seconds. ‘action=swipe’: Swipe from the starting point with coordinate (x, y) to the end point with coordinates2 (x2, y2). ‘action=type’: Input the specified text into the activated input box. ‘action=system\_button’: Press the system button. ‘action=open’: Open an app on the device. ‘action=wait’: Wait specified seconds for the change to happen. ‘action=terminate’: Terminate the current task and report its completion status.'',\\[2pt]

\hspace{1.5cm}``enum'': [``key'', ``click'', ``long\_press'', ``swipe'', ``type'', ``system\_button'', ``open'', ``wait'', ``terminate''],\\
\hspace{1.5cm}``type'': ``string''\\
\hspace{1.2cm}\},\\[6pt]

\hspace{1.2cm}``coordinate'': \{\\
\hspace{1.5cm}``description'': ``(x, y): The x (pixels from the left edge) and y (pixels from the top edge) coordinates to move the mouse to. Required only by ‘action=click’, ‘action=long\_press’, and ‘action=swipe’.'',\\
\hspace{1.5cm}``type'': ``array''\\
\hspace{1.2cm}\},\\[6pt]

\hspace{1.2cm}``coordinate2'': \{\\
\hspace{1.5cm}``description'': ``(x, y): Required only by ‘action=swipe’.'',\\
\hspace{1.5cm}``type'': ``array''\\
\hspace{1.2cm}\},\\[6pt]

\hspace{1.2cm}``text'': \{\\
\hspace{1.5cm}``description'': ``Required only by ‘action=key’, ‘action=type’, and ‘action=open’.'',\\
\hspace{1.5cm}``type'': ``string''\\
\hspace{1.2cm}\},\\[6pt]

\hspace{1.2cm}``time'': \{\\
\hspace{1.5cm}``description'': ``The seconds to wait. Required only by ‘action=long\_press’ and ‘action=wait’.'',\\
\hspace{1.5cm}``type'': ``number''\\
\hspace{1.2cm}\},\\[6pt]

\hspace{1.2cm}``button'': \{\\
\hspace{1.5cm}``description'': ``Back means returning to the previous interface, Home means returning to the desktop, Menu means opening the application background menu, and Enter means pressing the enter. Required only by ‘action=system\_button’'',\\
\hspace{1.5cm}``enum'': [``Back'', ``Home'', ``Menu'', ``Enter''],\\
\hspace{1.5cm}``type'': ``string''\\
\hspace{1.2cm}\},\\[6pt]

\hspace{1.2cm}``status'': \{\\
\hspace{1.5cm}``description'': ``The status of the task. Required only by ‘action=terminate’.'',\\
\hspace{1.5cm}``type'': ``string'',\\
\hspace{1.5cm}``enum'': [``success'', ``failure'']\\
\hspace{1.2cm}\}\\

\hspace{0.9cm}\},\\[4pt]

\hspace{0.9cm}``required'': [``action''],\\
\hspace{0.9cm}``type'': ``object''\\

\hspace{0.6cm}\},\\[4pt]

\hspace{0.6cm}``args\_format'': ``Format the arguments as a JSON object.''\\

\hspace{0.3cm}\}\\

\}\\ 

\medskip

\textbf{Note:}\\
For the ‘swipe’ action, pay attention to direction. Screen coordinates use a top-left origin, so the change in coordinates follows the finger’s drag, while the page/content moves in the opposite direction.\\[2pt]

For reference:\\
\hspace{0.3cm}- If y1 $\textgreater$ y2 $\rightarrow$ swipe up.\\
\hspace{0.3cm}- If y1 $\textless$ y2 $\rightarrow$ swipe down.\\
\hspace{0.3cm}- If x1 $\textgreater$ x2 $\rightarrow$ swipe left.\\
\hspace{0.3cm}- If x1 $\textless$ x2 $\rightarrow$ swipe right.\\[4pt]

When both axes change, determine the primary direction by comparing \(|\Delta x|\) and \(|\Delta y|\).

\medskip

\textbf{Task Instruction:}\\
\{TASK\_INSTRUCTION\}

\medskip

\textbf{Action History:}\\
\{ACTION\_HISTORY\}

\medskip

\textbf{Current Action:}\\
\{ACTION\}

\medskip

Now please generate critiques and evaluate correctness.

\end{longtable}

\twocolumn
\onecolumn

\begin{longtable}{p{0.96\textwidth}}
\caption{Unified desktop \& web critic prompt}
\label{tab:critic_prompt_desktop} \\
\toprule
\textbf{Desktop \& Web Critic Prompt} \\
\midrule
\endfirsthead

\toprule
\textbf{Desktop \& Web Critic Prompt} \\
\midrule
\endhead

\bottomrule
\endfoot

You are an expert GUI task evaluator. Your role is to analyze the provided action in context, generate an insightful textual critique, and grade the action’s effectiveness.

Your evaluation must be based on the current screen image, the overall task instruction, and the history of previous actions.
At the end of your critique, you must provide a final grade in this exact format:
``Verification: Does this action contribute to the completion of the task? (Yes/No)~X'', where X is either Yes or No.

\medskip
\textbf{Action Space Specification:}

\{

\hspace{0.3cm}``type'': ``function'',\\[2pt]

\hspace{0.3cm}``function'': \{\\

\hspace{0.6cm}``name\_for\_human'': ``computer\_use'',\\
\hspace{0.6cm}``name'': ``computer\_use'',\\[2pt]

\hspace{0.6cm}``description'': ``Use a mouse and keyboard to interact with a computer, and take screenshots. This is an interface to a desktop GUI. You do not have access to a terminal or applications menu. You must click on desktop icons to start applications. Some applications may take time to start or process actions, so you may need to wait and take successive screenshots to see the results of your actions. The screen’s resolution is \{SCREEN\_WIDTH\}x\{SCREEN\_HEIGHT\}. Whenever you intend to move the cursor to click on an element like an icon, you should consult a screenshot to determine the coordinates of the element before moving the cursor. If you tried clicking on a program or link but it failed to load, even after waiting, try adjusting your cursor position so that the tip of the cursor visually falls on the element that you want to click. Make sure to click any buttons, links, icons, etc with the cursor tip in the center of the element. Don’t click boxes on their edges.'',\\[4pt]

\hspace{0.6cm}``parameters'': \{\\

\hspace{0.9cm}``properties'': \{\\

\hspace{1.2cm}``action'': \{\\
\hspace{1.5cm}``description'': ``The action to perform. The available actions are: ‘action=key’: Performs key-down presses on the arguments passed in order, then performs key releases in reverse order. ‘action=type’: Type a string of text on the keyboard. ‘action=mouse\_move’: Move the cursor to a specified (x, y) pixel coordinate on the screen. ‘action=left\_click’: Click the left mouse button at a specified (x, y) pixel coordinate on the screen. ‘action=left\_click\_drag’: Click and drag the cursor to a specified (x, y) pixel coordinate on the screen. ‘action=right\_click’: Click the right mouse button. ‘action=middle\_click’: Click the middle mouse button. ‘action=double\_click’: Double-click the left mouse button. ‘action=scroll’: Performs a scroll of the mouse wheel (positive = up, negative = down). ‘action=wait’: Wait specified seconds. ‘action=terminate’: Terminate the current task and report its completion status.'',\\[2pt]

\hspace{1.5cm}``enum'': [``key'', ``type'', ``mouse\_move'', ``left\_click'', ``left\_click\_drag'', ``right\_click'', ``middle\_click'', ``double\_click'', ``scroll'', ``wait'', ``terminate''],\\
\hspace{1.5cm}``type'': ``string''\\
\hspace{1.2cm}\},\\[6pt]

\hspace{1.2cm}``keys'': \{\\
\hspace{1.5cm}``description'': ``Required only by ‘action=key’.'',\\
\hspace{1.5cm}``type'': ``array''\\
\hspace{1.2cm}\},\\[6pt]

\hspace{1.2cm}``text'': \{\\
\hspace{1.5cm}``description'': ``Required only by ‘action=type’.'',\\
\hspace{1.5cm}``type'': ``string''\\
\hspace{1.2cm}\},\\[6pt]

\hspace{1.2cm}``coordinate'': \{\\
\hspace{1.5cm}``description'': ``(x, y): Required for pointer actions such as ‘action=left\_click’, ‘action=mouse\_move’, ‘action=left\_click\_drag’, etc.'',\\
\hspace{1.5cm}``type'': ``array''\\
\hspace{1.2cm}\},\\[6pt]

\hspace{1.2cm}``pixels'': \{\\
\hspace{1.5cm}``description'': ``The amount of scrolling to perform. Positive values scroll up, negative values scroll down. Required only by ‘action=scroll’.'',\\
\hspace{1.5cm}``type'': ``number''\\
\hspace{1.2cm}\},\\[6pt]

\hspace{1.2cm}``time'': \{\\
\hspace{1.5cm}``description'': ``The seconds to wait. Required only by ‘action=wait’.'',\\
\hspace{1.5cm}``type'': ``number''\\
\hspace{1.2cm}\},\\[6pt]

\hspace{1.2cm}``status'': \{\\
\hspace{1.5cm}``description'': ``The status of the task. Required only by ‘action=terminate’.'',\\
\hspace{1.5cm}``type'': ``string'',\\
\hspace{1.5cm}``enum'': [``success'', ``failure'']\\
\hspace{1.2cm}\}\\

\hspace{0.9cm}\},\\[4pt]

\hspace{0.9cm}``required'': [``action''],\\
\hspace{0.9cm}``type'': ``object''\\

\hspace{0.6cm}\},\\[4pt]

\hspace{0.6cm}``args\_format'': ``Format the arguments as a JSON object.''\\

\hspace{0.3cm}\}\\

\}\\ 

\medskip

\textbf{Task Instruction:}\\
\{TASK\_INSTRUCTION\}

\medskip

\textbf{Action History:}\\
\{ACTION\_HISTORY\}

\medskip

\textbf{Current Action:}\\
\{ACTION\}

\medskip

Now please generate critiques and evaluate correctness.

\end{longtable}

\twocolumn


\end{document}